\documentclass[letterpaper, 10 pt, conference]{ieeeconf}  

\usepackage{xifthen}
\usepackage{xparse}
\usepackage{subdepth}
\usepackage{leftidx}
\usepackage{bm}

\newcommand{\bbm}{\begin{bmatrix}}
\newcommand{\ebm}{\end{bmatrix}}

\DeclareMathAlphabet{\mbf}{OT1}{ptm}{b}{n}
\newcommand{\mbs}[1]{{\bm{#1}}}

\newcommand{\mbsbar}[1]{{\overline{\boldsymbol{#1}}}}
\newcommand{\mbshat}[1]{{\hat{\boldsymbol{#1}}}}
\newcommand{\mbstilde}[1]{{\tilde{\boldsymbol{#1}}}}

\newcommand{\mbsdot}[1]{{\dot {\boldsymbol{#1}}}}

\newcommand{\mbfbar}[1]{{\overline{\mbf{#1}}}}
\newcommand{\mbfhat}[1]{{\hat{\mbf{#1}}}}
\newcommand{\mbftilde}[1]{{\tilde{\mbf{#1}}}}

\newcommand{\mbfdot}[1]{{\dot{\mbf{#1}}}}

\newcommand{\cframe}[1]{{\smash{\protect\underrightarrow{\mathcal{F}}_{#1}}}}

\DeclareMathAlphabet{\mathbfit}{OML}{cmm}{b}{it}
\newcommand{\homo}[1]{{\mathbfit{#1}}}

\newcommand{\mbfh}[1]{{\homo{#1}}}

\newcommand{\pos}[2]{\leftidx{_{#1}}{ \mbf r}{_{#2}}} 

\newcommand{\vel}[3]{\leftidx{_{#1}}{\mbf v}{\IfValueTF{#2}{_{#2#3\hspace{2pt}}}{}}} 
\newcommand{\veltilde}[3]{\leftidx{_{#1}}{\mbftilde v}{\IfValueTF{#2}{_{#2#3\hspace{2pt}}}{}}} 
\newcommand{\velbar}[3]{\leftidx{_{#1}}{\mbfbar v}{\IfValueTF{#2}{_{#2#3\hspace{2pt}}}{}}} 
\newcommand{\velhat}[3]{\leftidx{_{#1}}{\mbfhat v}{\IfValueTF{#2}{_{#2#3\hspace{2pt}}}{}}} 
\newcommand{\veldot}[3]{\leftidx{_{#1}}{\mbfdot v}{\IfValueTF{#2}{_{#2#3\hspace{2pt}}}{}}} 

\newcommand{\myvec}[2]{\leftidx{_{#2}\hspace{-1pt}}{\mbf #1}{}} 

\newcommand{\acc}[3]{\leftidx{_{#1}}{\mbf a}{\IfValueTF{#2}{_{#2#3\hspace{2pt}}}{}}} 
\newcommand{\acctilde}[3]{\leftidx{_{#1}}{\mbftilde a}{\IfValueTF{#2}{_{#2#3\hspace{2pt}}}{}}} 
\newcommand{\accbar}[3]{\leftidx{_{#1}}{\mbfbar a}{\IfValueTF{#2}{_{#2#3\hspace{2pt}}}{}}} 

\newcommand{\rotvel}[3]{\leftidx{_{#1}}{\mbs \omega}{\IfValueTF{#2}{_{#2#3\hspace{2pt}}}{}}} 
\newcommand{\rotveltilde}[3]{\leftidx{_{#1}}{\mbstilde \omega}{\IfValueTF{#2}{_{#2#3\hspace{2pt}}}{}}} 
\newcommand{\rotvelbar}[3]{\leftidx{_{#1}}{\mbsbar \omega}{\IfValueTF{#2}{_{#2#3\hspace{2pt}}}{}}} 
\newcommand{\rotvelhat}[3]{\leftidx{_{#1}}{\mbshat \omega}{\IfValueTF{#2}{_{#2#3\hspace{2pt}}}{}}} 
\newcommand{\rotveldot}[3]{\leftidx{_{#1}}{\mbsdot \omega}{\IfValueTF{#2}{_{#2#3\hspace{2pt}}}{}}} 

\newcommand{\T}[2]{\leftidx{}{\mbfh T}{_{#1#2\hspace{2pt}}}} 
\newcommand{\q}[2]{\leftidx{}{\mbf q}{_{#1#2\hspace{2pt}}}} 

\IEEEoverridecommandlockouts                              

\usepackage{graphics} 
\usepackage{epsfig} 
\usepackage{amsmath} 
\usepackage{amssymb}  
\usepackage{amsfonts}
\usepackage{cite}
\usepackage{color}
\usepackage{multirow}
\usepackage{array}
\usepackage{subfigure}
\usepackage{caption}
\usepackage{booktabs}
\usepackage[keeplastbox]{flushend}
\usepackage{hyperref}

\title{\LARGE \bf
Dense RGB-D-Inertial SLAM with Map Deformations
}

\author{Tristan Laidlow, Michael Bloesch, Wenbin Li and Stefan Leutenegger
\thanks{The authors are with the Dyson Robotics Laboratory,
        Imperial College London, UK. Corresponding author: Tristan Laidlow, {\tt\small t.laidlow15@imperial.ac.uk}}%
\thanks{Research presented in this paper has been supported by Dyson Technology Ltd.}%
}

\newcommand\ts{\rule{0pt}{2.8ex}}       

\begin{document}

\maketitle
\thispagestyle{empty}
\pagestyle{empty}


\begin{abstract}

While dense visual SLAM methods are capable of estimating dense reconstructions of the environment, they suffer from a lack of robustness in their tracking step, especially when the optimisation is poorly initialised.
Sparse visual SLAM systems have attained high levels of accuracy and robustness through the inclusion of inertial measurements in a tightly-coupled fusion.
Inspired by this performance, we propose the first tightly-coupled dense RGB-D-inertial SLAM system.

Our system has real-time capability while running on a GPU.
It jointly optimises for the camera pose, velocity, IMU biases and gravity direction while building up a globally consistent, fully dense surfel-based 3D reconstruction of the environment.
Through a series of experiments on both synthetic and real world datasets, we show that our dense visual-inertial SLAM system is more robust to fast motions and periods of low texture and low geometric variation than a related RGB-D-only SLAM system.

\end{abstract}


\section{INTRODUCTION}

Visual Simultaneous Localisation and Mapping (SLAM) has achieved a level of maturity that allows for integration into mobile robots.
We can split respective algorithms into two broad categories: sparse landmark-based systems and dense or semi-dense systems.
While sparse methods may not directly produce a map that is useful for robot navigation, pose estimation quality and robustness of state-of-the-art systems, such as \cite{Mur-Artal:etal:TRO2015} and \cite{Forster:etal:ICRA2014}, are typically very high.
Even higher accuracy and robustness may be attained by the inclusion of inertial measurements in a tightly-coupled fusion. Inertial Measurement Units (IMUs) have become very cheap and are abundant in today's consumer electronic devices, therefore their use in visual SLAM has been widely adopted. Approaches are formulated either as filters e.g.\ \cite{Mourikis:Roumeliotis:ICRA2007, Li:Mourikis:2013, Tanskanen:etal:IROS2015, Bloesch:etal:IROS2015} or as methods employing iterative minimisation, typically in a sliding window manner, such as \cite{Jones:Soatto:IJRR2011, Keivan:etal:ISER2014, Leutenegger:etal:IJRR2014, Forster:etal:RSS2015}. Loosely-coupled approaches to visual-inertial fusion, such as \cite{Weiss:etal:ICRA2012, Engel:etal:IROS2012, Meier:etal:ICRA2011} that separate out either the visual or inertial estimation part have also been proposed. These methods are popular due to their modularity, but disregard correlations in the state estimates, typically leading to lower accuracy and/or robustness.

Other research has focused on producing denser maps, a development enabled by ever more computational power and specifically the emergence of Graphics Processing Units (GPUs), as well as by novel sensors in the form of depth cameras (RGB-D cameras).
Such systems typically employ direct photometric alignment of the image and/or Iterative Closest Point (ICP) alignment of the depth image to the map in a \emph{tracking} step; in a separate \emph{mapping} step, new information is fused into the dense map representation. 
Examples using solely monocular cameras range from the fully dense DTAM \cite{Newcombe:etal:ICCV2011} to LSD-SLAM \cite{Engel:etal:ECCV2014}, which reconstructs a semi-dense representation.
RGB-D SLAM approaches that make use of depth cameras include KinectFusion \cite{Newcombe:etal:ISMAR2011} and other methods employing signed distance function-based volumetric mapping, such as \cite{Kerl:etal:IROS2013,Whelan:etal:RSSRGBD2012,Kahler:etal:ISMAR2015}. Surfel-based mapping presents itself as an alternative enabling easier scalability in space and time: ElasticFusion \cite{Whelan:etal:IJRR2016}, which our proposed work is based on, focuses on global map consistency by applying elastic map deformations upon loop closure. 

\begin{figure}[t]
\centering
\includegraphics[width=1\linewidth]{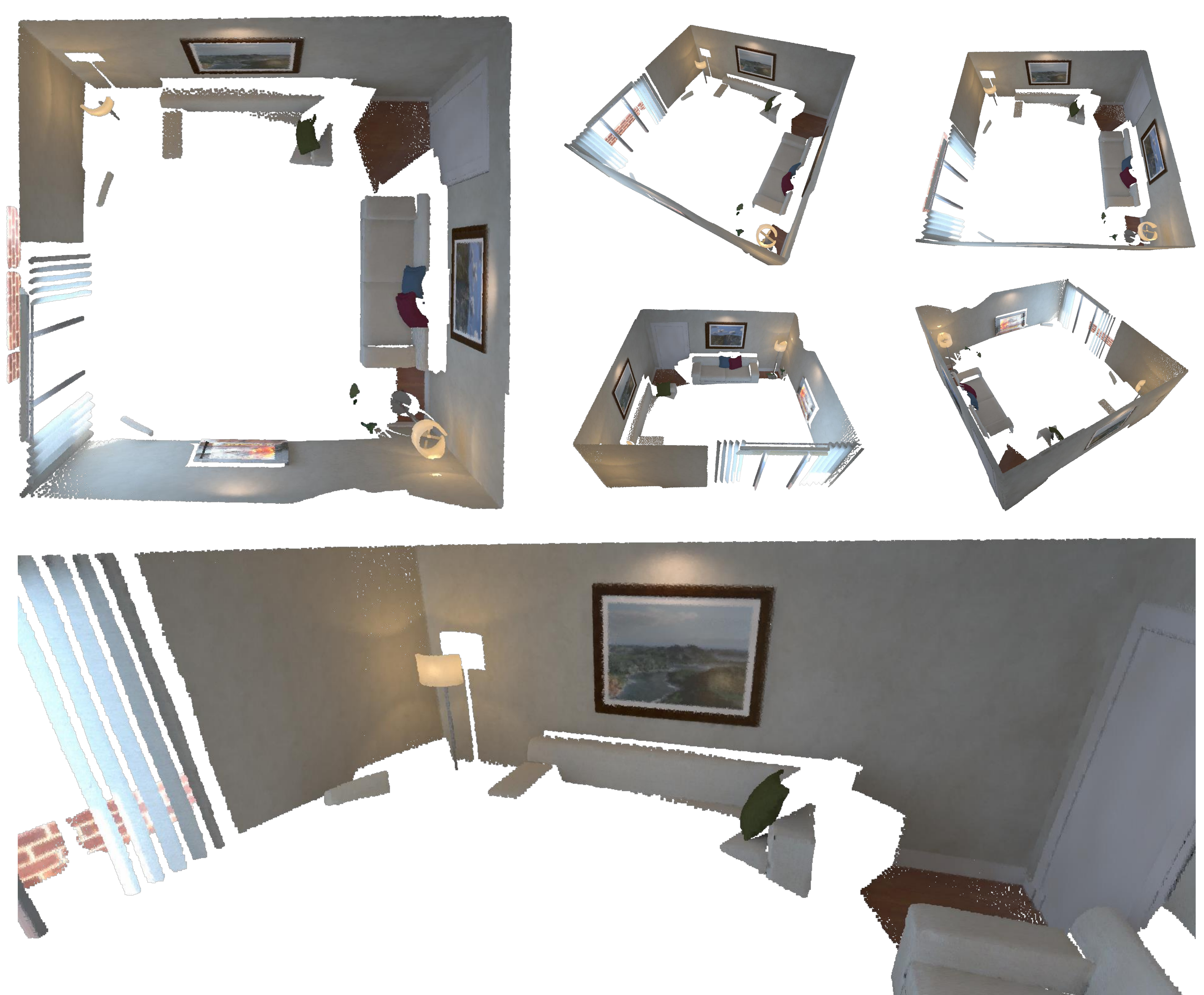}
\caption{Tightly integrating IMU measurements into a dense RGB-D SLAM system leads to more accurate and robust tracking and 3D reconstruction compared with using visual information alone. The above surface reconstruction was captured despite periods of low texture and geometric variation.}
\label{fig:teaser}

\vspace{-5mm}

\end{figure}

Dense maps offer much more potential for safe robot navigation which we envisage to evolve into very general spatial perception with semantic understanding and tracking of dynamic objects in the future.
As of now, however, vision-only SLAM, and dense SLAM using direct image alignment in particular, suffers from a lack of robustness in the tracking step when initialised too far from the ``true'' solution; in fact, the tracking optimisation may not converge at all in absence of sufficient texture and/or geometric variation in the depth channel.
To address these shortcomings, inspired by the success of sparse visual-inertial systems, we advocate the integration of acceleration and rotation rate measurements into the tracking of a dense SLAM system.
In principle, the tight integration of these complementary sensing modalities should provide robustness in rapid motion, low texture and flat walls.
Furthermore, the inclusion of an IMU renders the gravity direction observable, which not only improves map accuracy due to bounded absolute inclination error, but may also be of paramount importance for robot control, most prominently drones.

There have been a few recent examples of dense visual-inertial systems: both \cite{Omari:etal:ICRA2015} and \cite{Ma:etal:ISER2015} present loosely-coupled approaches, with the former using the integrated IMU data as a prediction step in a filter to estimate the transformation between image pairs, and the latter fusing relative poses generated by inertial and stereo camera measurements in a manner similar to a pose graph.
A tightly-coupled semi-dense monocular visual-inertial odometry system is presented in \cite{Concha:etal:ICRA2016}. Unlike other pure monocular odometry systems, it is able to use the inertial data to remove scale ambiguity.
Their system uses a semi-dense approach for tracking and, in a separate thread, estimates a fully dense map below frame rate using a piecewise planar prior.
Another example of a semi-dense visual-inertial odometry system is described in  \cite{Usenko:etal:ICRA2016}.
This system is implemented within the stereo LSD-SLAM framework \cite{Engel:etal:IROS2015}.
Through a series of experiments, they demonstrate that their tightly-coupled approach outperforms both vision-only and loosely-coupled approaches. While the system is closely related to ours, we propose a more map-centric, fully dense approach that additionally considers a depth channel and performs map optimisation compliant with gravity alignment.

In this paper, we extend the RGB-D SLAM system ElasticFusion \cite{Whelan:etal:IJRR2016} with tightly-coupled IMU integration, which is capable of more accurate and robust fully dense mapping. Please see Figure \ref{fig:teaser} for an example map output. More specifically, we make the following contributions:
\begin{itemize}
\item In the tracking step, we simultaneously estimate the camera pose, velocity,
IMU biases and gravity direction from an RGB-D camera and IMU by minimising a joint photometric, geometric, and inertial energy functional.
\item Concerning the mapping, we propose a system that constructs a globally consistent, fully dense surfel-based 3D reconstruction of the environment.
The map is optimised not through a pose graph, but by applying non-rigid space deformations using a sparse deformation graph.
We propose an addition to the deformation energy that ensures consistency with the observable gravity direction.
\item Through experiments on both synthetic and real world datasets, we demonstrate
the benefits of our approach.
It performs well under aggressive motion, fast rotations, and under low texture and geometric variation.
We demonstrate trajectory and map reconstruction accuracy higher or on-par with an RGB-D-only ElasticFusion.
\item We emphasise that the system maintains real-time capability while running on a GPU.
Unlike \cite{Concha:etal:ICRA2016} and \cite{Usenko:etal:ICRA2016}, which achieve real-time performance on a CPU, our system constructs a fully dense map at frame rate.
\item To the best of our knowledge, we hereby present the first tightly-coupled dense RGB-D-inertial SLAM system.
\end{itemize}

The remainder of this paper is organised as follows: we start with an overview of notation employed in Section \ref{s:notation} and of the approach as such in \ref{s:overview}.
We then describe the method concerning tracking in \ref{s:tracking} and mapping in \ref{s:mapping} followed by extensive results in \ref{s:results}.


\section{NOTATION}
\label{s:notation}
Throughout this work we will employ the following notation: a reference frame $A$ is denoted $\cframe{A}$, with vectors expressed in it denoted as $\myvec{p}{A}$.
The position vector from the origin of $\cframe{A}$ to the origin of $\cframe{B}$, represented in  $\cframe{A}$ is written $\pos{A}{B}$ and the velocity of the origin of $\cframe{C}$ as observed by $\cframe{B}$ and expressed in $\cframe{A}$ is denoted $\vel{A}{B}{C}$.
The homogeneous transformation matrix that transforms homogeneous points from $\cframe{B}$ to $\cframe{A}$ is written as $\T{A}{B}$.
The corresponding rotation is represented by a Hamiltonian unit quaternion, $\q{A}{B}$.
In order to refer to the homogeneous coordinates of a coordinate vector $\mbf p$ we will use the italic notation $\homo p$.

Four different coordinate frames will be used in this work:
\begin{itemize}
\item $\cframe{W}$, the world frame in which the global model is expressed.
This frame corresponds with the initial camera frame.
\item $\cframe{I}$, the inertial frame that is aligned with gravity and shares an origin with $\cframe{W}$.
\item $\cframe{C}$, the camera frame in which the RGB-D data is observed.
\item $\cframe{S}$, the sensor frame in which the IMU data is observed.
\end{itemize}


\begin{figure}[t]
    \centerline{
    \subfigure[RGB-D-only ElasticFusion]{\includegraphics[height=0.9in]{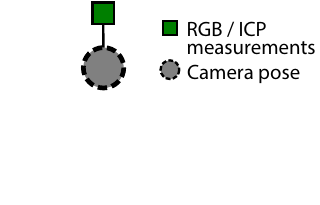}}
    ~~~~
    \subfigure[RGB-D-inertial ElasticFusion]{\includegraphics[height=0.9in]{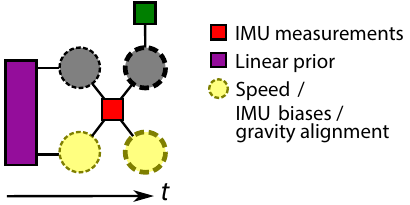}}
    }
    \caption{Tracking optimisation in RGB-D-only vs.\ RGB-D-inertial ElasticFusion: inertial measurements necessitate the augmentation of the state with speed, IMU biases, and gravity alignment; furthermore, the temporal nature of IMU measurements requires us to marginalise old states, resulting in a linear prior. }
    \label{fig:sys-rgbdi}
    
    \vspace{-5mm}
    
\end{figure}

\section{SYSTEM OVERVIEW}
\label{s:overview}
Our system directly builds upon the vision-only dense RGB-D tracking and mapping approach of ElasticFusion \cite{Whelan:etal:IJRR2016}.
Like ElasticFusion, our approach performs the tracking and mapping in separate steps.
In the tracking step, a joint photometric, geometric and inertial energy functional is constructed.
Please see Figure \ref{fig:sys-rgbdi} for a comparison of the factor-graph representation of the underlying tracking optimisation problem in RGB-D-only and RGB-D-inertial ElasticFusion.
Whereas ElasticFusion combined the photometric and geometric terms based on a tuning parameter $\lambda$, we combine the terms based on the covariances associated with the measurement noise.
A nonlinear optimisation formulation is then used to simultaneously estimate the camera pose, velocity, IMU biases and gravity direction.
Unlike the original ElasticFusion which only estimated the current camera pose, our system estimates the states associated with both the current and previous camera frames.
After the optimisation the states related to the previous frame are marginalised and the remaining current state is used as a prior for the next time step.

In the mapping step, a fully dense surfel-based surface representation is constructed from the camera data and estimated poses obtained from the tracking step.
The map is kept globally consistent by applying non-rigid space deformations through a sparse deformation graph.
We extend the deformation energy formulation proposed by ElasticFusion to ensure consistency with the observable gravity direction.

\section{TRACKING}
\label{s:tracking}
\subsection{States \& Local Parameterisation}

At the arrival of each new camera frame, we estimate the current state, $\mbf x_1$, while simultaneously refining the previous state, $\mbf x_0$.
The system state is comprised of the camera position in the world frame $\pos{W}{C}$, the camera orientation $\q{W}{C}$, the velocity of the IMU in the inertial frame $\vel{I}{I}{S}$, the biases of the gyroscopes $\mbf b_{g}$ and accelerometers $\mbf b_{a}$, and the orientation of the world frame in the inertial frame $\q{I}{W}$.
Therefore the system state $\mbf x$ for a specific time instance is given by:
\begin{align}
\mbf x &:= \left[ \pos{W}{C}^{T}, \q{W}{C}^{T}, \vel{I}{I}{S}^{T}, \mbf b_{g}^{T}, \mbf b_{a}^{T}, \q{I}{W}^{T}  \right]^T \\
& \ \ \ \ \in \mathbb{R}^{3} \times S^{3} \times \mathbb{R}^{9} \times S^{3}. \nonumber
\end{align}
While only two degrees of freedom are required to express the gravity direction in the world frame, for simplicity, we use a 3D implementation with gauge freedom.
We did not observe any issues related to this formulation.

The system state exists on a manifold and so is updated by a local perturbation $\delta \mbf x$ in the tangent space through the $\boxplus$ operator, such that $\mbf x = \bar{\mbf x} \boxplus \delta \mbf x$ around a reference $\bar{\mbf x}$.
For $\pos{W}{C}$, $\vel{I}{I}{S}$, $\mbf b_{g}$ and $\mbf b_{a}$, the $\boxplus$ operator is equivalent to standard vector addition. For $\q{W}{C}$ and $\q{I}{W}$, a combination of the group operator (quaternion multiplication) and exponential map is used ($\mbf q \boxplus \delta \mbs \alpha = \exp(\delta \mbs \alpha) \otimes \mbf q$).
This results in the following minimal local coordinate representation:
\begin{equation}
\delta \mbs x = \left[ \delta \mbf r^{T}, \delta \mbs \alpha^{T}, \delta \mbf v^{T}, \delta \mbf b_{g}^{T}, \delta \mbf b_{a}^{T}, \delta \mbf g^{T} \right]^T \in \mathbb{R}^{18}.
\end{equation}
Similarly, a $\boxminus$ operator can be introduced to compute the difference between two systems states.
For regular vector space quantities this corresponds to standard subtraction.
For orientations an inverse of the above $\boxplus$ can be constructed ($\mbf p \boxminus \mbf q = \exp^{-1}(\mbf p \otimes \mbf q^{-1})$).

Please refer to \cite{Hertzberg:etal:INFORMATIONFUSION2011,Bloesch:etal:CoRR2016} for further details.

\subsection{Dense Photometric \& Geometric Alignment}

The RGB-D subsystem of our approach combines dense per-pixel photometric alignment with ICP point-to-plane geometric alignment.
The photometric alignment error, $e_{\mathrm{RGB}, \mbf u}$ for pixel $\mbf u$ in the current image is the intensity difference between the transformed previous and current images:
\begin{align}
e_{\mathrm{RGB}, \mbf u} =& \ I_0\left( \mbs \pi \left( \mbf K \, \T{W}{C_{0}}^{-1} \T{W}{C_{1}} \mbs \rho(\mbs u, d) \right) \right) - I_1(\mbs u),
\end{align}
where $I_{*}(\cdot)$ is a scalar function that returns the intensity value of a given pixel, $\mbs \pi(\cdot)$ is the projection and dehomogenisation function that maps a 3D point onto the image plane, and $\mbs \rho(\mbf u, d)$ is the back-projection function that returns a homogeneous 3D point for pixel $\mbf u$ with a depth $d$. $\mbf K$ is the camera intrinsics matrix, containing the focal lengths and principal point of the camera.

The geometric alignment error uses a point-to-plane ICP technique and computes the signed distance between a point $\mbf p_{k}$ projected from the depth measurement $k$ as viewed from the current camera pose and a corresponding point in the global model:
\begin{equation}
e_{\mathrm{ICP}, k} = \myvec{n}{W}_{k}^{T} \left[ \T{W}{C_{1}} {}_{C_{1}}\homo{p}_{k} - {}_{W}\homo{p}_{k} \right]_{1:3}.
\end{equation}

\subsection{Inertial Integration}

For the formulation of the IMU measurement error term we adopt the approach of \cite{Leutenegger:etal:IJRR2014}, extending it to include the preintegration technique described by \cite{Forster:etal:RSS2015}.
The IMU measurements are integrated numerically between the previous and current camera frames.
The final IMU error term is given by:
\begin{equation}
\mbf e_\mathrm{IMU} = \hat{\mbf x}_{1}(\mbf x_{0}) \boxminus \mbf x_{1},
\end{equation}
where $\hat{\mbf x}_{1}$ is the prediction of the current state by integrating the IMU measurements onto the previous state.

\subsection{Optimisation}

The RGB-D-inertial tracking problem is solved using a joint cost function $c_{\mathrm{track}}$ that contains the weighted photometric alignment, geometric alignment and inertial terms:
\begin{align}
c_\mathrm{track}&(\mbf x_{0}, \mbf x_{1}) = \sum_{\mbf u} e_{\mathrm{RGB},\mbf u} W_\mathrm{RGB} e_{\mathrm{RGB}, \mbf u} \\
 &+ \sum_{k} e_{\mathrm{ICP}, k} W_{\mathrm{ICP}, k} e_{\mathrm{ICP}, k} + \mbf e_\mathrm{IMU}^{T} \mbf W_\mathrm{IMU} \mbf e_\mathrm{IMU} \nonumber \\
 &+ \left(\mbf x_{0} \boxminus \bar{\mbf x}_{0} - \mbf {H^{*}}^{-1} \mbf b^{*} \right)^{T} \mbf {H^{*}} \left( \mbf x_{0} \boxminus \bar{\mbf x}_{0} - \mbf {H^{*}}^{-1} \mbf b^{*} \right), \nonumber
\end{align}
where $W_\mathrm{RGB}$, $W_{\mathrm{ICP}, k}$, and $\mbf W_\mathrm{IMU}$ are the inverse covariance (matrices) associated with the respective measurement uncertainties, and $\mbf H^{*}$ and $\mbf b^{*}$ are priors obtained through the marginalisation step.

The cost function is minimised using a Gauss-Newton iterative method with a three level coarse-to-fine pyramid scheme.
We omit the Jacobians for readability and space constraints.
After each iteration, the current and previous states are updated using the $\boxplus$ operator.

\subsection{Partial Marginalisation \& Fixation of Variables}

The equations for the Gauss-Newton system are constructed from the Jacobians, error terms and information relating to the current and previous states, taking the form:
\begin{equation}
\left[ \begin{array}{c c} \mbf H_{00} & \mbf H_{01} \\ \mbf H_{10} & \mbf H_{11} \end{array} \right]
\left[ \begin{array}{c} \delta \mbf x_{0} \\ \delta \mbf x_{1} \end{array} \right] = 
\left[ \begin{array}{c} \mbf b_{0} \\ \mbf b_{1} \end{array} \right].
\end{equation}

After the current and previous states are updated, we marginalise out the previous state using the Schur-Complement:
\begin{subequations}
\begin{align}
\mbf H_{11}^{*} &= \mbf H_{11} - \mbf H_{10} \mbf H_{00}^{-1} \mbf H_{01}, \\
\mbf b_{1}^{*} &= \mbf b_{1} - \mbf H_{10} \mbf H_{00}^{-1} \mbf b_{0}.
\end{align}
\end{subequations}

The resulting $\mbf H^{*}$ and $\mbf b^{*}$ information is used as a prior in the next optimisation step.
This partial marginalisation fixes the linearisation point, but with each iteration in the subsequent optimisation scheme, the linearisation point changes.
Instead of relinearising at each step, we apply a first-order correction, $\Delta \mbf x$, based on the difference between the new and old linearisation points as is commonly done in the literature \cite{Leutenegger:etal:IJRR2014}\cite{Usenko:etal:ICRA2016}:
\begin{subequations}
\begin{align}
\mbf H_{11}^{*'} &= \mbf H_{11}^{*}, \\
\mbf b_{1}^{*'} &= \mbf b_{1}^{*} + \mbf H_{11}^{*}\Delta \mbf x.
\end{align}
\end{subequations}


\section{MAPPING}
\label{s:mapping}
Like the original ElasticFusion, the map in our formulation is split into \emph{active} and \emph{inactive} areas \cite{Whelan:etal:IJRR2016}.
The active map is the area most recently observed and is where the tracking and fusing takes place.
If a segment of the active map is not observed for a period of time, $\delta_{t}$, it becomes inactive.
We keep the map globally consistent by attempting to match the currently observed portion of the active map with the inactive map.
If a match is detected, the loop is closed by applying non-rigid space deformations through a sparse deformation graph.
A deformation graph is a set of nodes, $\mathcal{G}^{l}$, that are embedded in the global model, each with a position, $\mathcal{G}_{\mbf g}^{l}$, and a set of neighboring nodes, $\mathcal{G}^{n} \in \mathcal{N}(\mathcal{G}^{l})$.
Each deformation node stores a Euclidean transformation as a rotation, $\mathcal{G}_{\mbf R}^{l}$, and a translation, $\mathcal{G}_{\mbf t}^{l}$, that is used to elastically deform surfels in the map from a source position $\mathcal{Q}_{\mbf s}$ to a destination position $\mathcal{Q}_{\mbf d}$ through a deformation function, $\phi(\cdot)$, defined in \cite{Whelan:etal:IJRR2016}.
This affine transformation is determined by minimising a cost function.
In the original ElasticFusion, the cost function consists of five terms. The first encourages rigidity in the deformation:
\begin{equation}
E_{rot} = \sum_{l} \left\lVert {\mathcal{G}_{\mbf R}^{l}}^{T}\mathcal{G}_{\mbf R}^{l} - \mbf I \right\rVert_{F}^{2}.
\end{equation}
The second encourages smoothness in the deformation:
\begin{equation}
\begin{split}
&E_\mathrm{reg} =\\ 
&\displaystyle\sum_{l} \sum_{n \in \mathcal{N}(\mathcal{G}^{l})} \left\lVert
\mathcal{G}_{\mbf R}^{l} (\mathcal{G}_{\mbf g}^{n} - \mathcal{G}_{\mbf g}^{l}) + \mathcal{G}_{\mbf g}^{l} + \mathcal{G}_{t}^{l} - (\mathcal{G}_{\mbf g}^{n} + \mathcal{G}_{t}^{n})
\right\rVert_{2}^{2}.
\end{split}
\end{equation}
The third minimises the distance of each point from the desired deformation:
\begin{equation}
E_\mathrm{con} = \sum_{p} \left\lVert \phi(\mathcal{Q}_{s}^{p}) - \mathcal{Q}_{d}^{p} \right\rVert_{2}^{2}.
\end{equation}
The fourth constrains the inactive areas of the map such that the active map is being deformed into the inactive map:
\begin{equation}
E_\mathrm{pin} = \sum_{p} \left\lVert \phi(\mathcal{Q}_{d}^{p}) - \mathcal{Q}_{d}^{p} \right\rVert_{2}^{2}.
\end{equation}
The fifth term is only applied to global deformations, and is used to prevent previous registrations, $\mathcal{R}$, from being pulled apart by future global loop closures:
\begin{equation}
E_\mathrm{rel} = \sum_{p} \left\lVert \phi(\mathcal{R}_{s}^{p}) - \phi(\mathcal{R}_{d}^{p}) \right\rVert_{2}^{2}.
\end{equation}

As matches between the active and inactive areas of the map are determined only by the RGB-D subsystem, we include a sixth cost term in our formulation to constrain the graph from deforming the map out of alignment with gravity:
\begin{equation}
E_\mathrm{imu} = \sum_{l} \left\lVert \mathcal{G}_{\mbf R}^{l} \myvec{g}{W} - \myvec{g}{W}. \right\rVert_{2}^{2},
\end{equation}
where ${_W}\mbf{g}$ denotes the acceleration due to gravity represented in vision-world frame $\cframe{W}$.

Keeping the parameter choices the same as ElasticFusion, the total cost function for local loop closures is given by:
\begin{equation}
E_\mathrm{loc} = \omega_{f} E_\mathrm{rot} + \omega_{r} E_\mathrm{reg} + \omega_{c}(E_\mathrm{con} + E_\mathrm{pin}) + \omega_{i} E_\mathrm{imu},
\end{equation}
and the total cost for the global loop closures is given by:
\begin{equation}
E_\mathrm{glo} = \omega_{f} E_\mathrm{rot} + \omega_{r} E_\mathrm{reg} + \omega_{c}(E_\mathrm{con} + E_\mathrm{pin} + E_\mathrm{rel}) + \omega_{i} E_\mathrm{imu},
\end{equation}
with $\omega_{f}$ = 1, $\omega_{r}$ = 10, and $\omega_{c}$ = $\omega_{i}$ = 100.


\section{RESULTS}
\label{s:results}
We evaluate our system in terms of trajectory estimation and reconstruction accuracy on both synthetic and real world datasets.
We adapt the living room sequences of the ICL-NUIM dataset \cite{Handa:etal:ICRA2014} for the experiments on synthetic data.
For the real world experiments, we recorded our own datasets along with ground truth poses from a Vicon motion capture system.
The synthetic dataset consists of slow, smooth trajectories usually required for dense visual SLAM.
The real world dataset contains a mixture of slow trajectories, aggressive motions and sequences with low texture and geometric information where vision-only systems tend to struggle.

We consider two metrics when examining the performance of the system: the absolute trajectory (ATE) root-mean-square error (RMSE) described in \cite{Sturm:etal:IROS2012}, and for reconstruction error, the mean distance from each point in the reconstruction to the nearest surface in the aligned ground truth model.
The ATE RMSE is calculated for all sequences, but the reconstruction error is only available for the synthetic dataset.
As the behavior of the loop closure mechanism is non-deterministic, we ran each test 10 times and took the average result.
Tests where either system had lost tracking are denoted by brackets in the tables.

Through these experiments, we show that our dense RGB-D-inertial SLAM system performs at least as well as the RGB-D-only system on ``easier'' trajectories where the problem is well constrained by the visual data alone, but is much more robust when facing sequences with fast motions or little photometric and geometric variation.

\begin{table}[h]
\caption{System Parameters}
\label{t:imu_params}
\begin{center}
\begin{tabular}{>{\arraybackslash}m{2.95cm} >{\centering\arraybackslash}m{1.0cm}>{\centering\arraybackslash}m{1.4cm} l}
\toprule
\textbf{Noise Parameter} & \textbf{Synthetic Dataset} & \textbf{Real World Dataset} & \textbf{Units} \\
\midrule
Gyr. saturation & 7.8 & 7.8 & rad s\textsuperscript{-1} \\
Acc.\ saturation & 176.0 & 176.0 & m s\textsuperscript{-2} \ts\\
Gyr. noise density & 12.0e-4 & 12.0e-4 & rad s\textsuperscript{-1} Hz\textsuperscript{-0.5} \ts\\
Acc.\ noise density & 8.0e-3 & 8.0e-2 & m s\textsuperscript{-2} Hz\textsuperscript{-0.5} \ts\\
Gyr. bias Prior & 0.03 & 0.03 & rad s\textsuperscript{-1} \ts\\
Acc.\ bias Prior & 0.1 & 1 & m s\textsuperscript{-2} \ts\\
Gyr. drift noise density & 4.0e-6 & 4.0e-6 & rad s\textsuperscript{-2} Hz\textsuperscript{-0.5} \ts\\
Acc.\ drift noise density & 2.0e-5 & 2.0e-5 &  m s\textsuperscript{-3} Hz\textsuperscript{-0.5} \ts\\
Acc. due to gravity& 9.81 & 9.81 & m s\textsuperscript{-2} \ts\\
IMU rate & 200 & 200 & Hz \ts\\
Static acc.\ bias & [0, 0, 0] & [0.060, 0.258, 0.126] & m s\textsuperscript{-2} \rule{0pt}{4.8ex}\\
Image intensity noise & 4.0 & 1.0 & - \ts\\
Image disparitiy noise & 5.5 & 5.5 & pixels \ts\\
\bottomrule
\end{tabular}
\end{center}
\end{table}

\subsection{Synthetic Data}

We evaluate both the trajectory estimation and surface reconstruction accuracy of our system on a modified version of the living room sequences in the ICL-NUIM dataset.
The ICL-NUIM dataset is a benchmark that provides ground truth poses as well as a 3D model with which to evaluate reconstructions of RGB-D SLAM systems.
The dataset does not come with inertial data, however, so in a manner similar to \cite{Kerl:etal:ICCV2015}, we fit splines to the ground truth poses to simulate continuous trajectories.
IMU measurements are then generated along these trajectories using the model described in \cite{Nikolic:etal:SENSORS2016} and the noise parameters given in Table \ref{t:imu_params}.
However, due to the non-smooth trajectories of the dataset, we needed to sample every 10th frame of the ground truth trajectories when fitting the splines.
This resulted in the new ground truth poses being close to but not exactly the same as those in the original dataset.
Therefore, we rendered the images at these new poses using POV-Ray and applied the same noise models to the images as those in the original dataset.

\begin{figure}[t]
\centering
\includegraphics[scale=0.53]{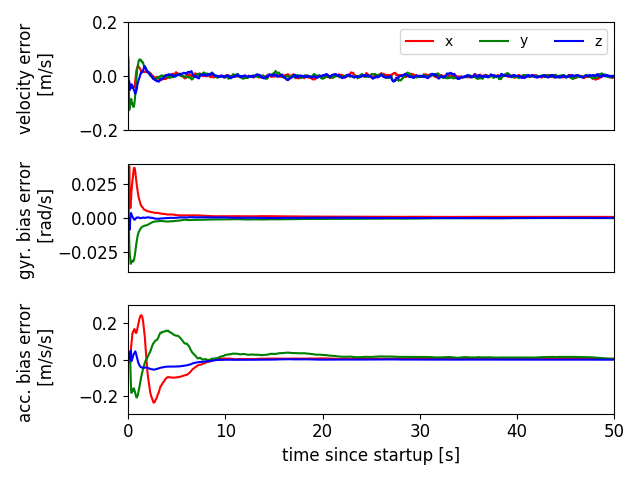}
\caption{Error in the velocity and bias estimates when compared to the ground truth values in synthetic dataset LR0. Our system is able to converge to and track the correct values.}
\label{fig:imuPlots}
\end{figure}

Since the entire ground truth states are known for the synthetic dataset, we are able to demonstrate that our system can accurately track the velocity and IMU biases.
For example, the error in the velocity and bias estimates for Sequence LR0, provided in Fig. \ref{fig:imuPlots}, quickly converges to zero.

We compared the performance of the RGB-D-inertial and RGB-D-only versions of our system on each of the four living room sequences of the modified ICL-NUIM dataset.
The results for the ATE RMSE are given in Table \ref{t:synth_ate} and for the reconstruction error in Table \ref{t:synth_recon}.

\begin{table}[h!]
\caption{ATE RMSE on the synthetic datasets (brackets indicate a tracking failure)}
\label{t:synth_ate}
\begin{center}
\begin{tabular}{lcc}
\toprule
\textbf{Sequence} & \textbf{RGB-D-Only} & \textbf{RGB-D-Inertial} \\
\midrule
LR0 & 0.032 & \bf{0.009} \\ 
LR1 & \bf{0.009} & 0.012 \\ 
LR2 & \bf{0.009} & \bf{0.009} \\ 
LR3 & (0.906) & \bf{0.019} \\ 
\bottomrule 
\end{tabular}
\end{center}
\end{table}

\begin{table}[h!]
\caption{Surface reconstruction accuracy on the synthetic datasets (brackets indicate a tracking failure)}
\label{t:synth_recon}
\begin{center}
\begin{tabular}{lcc}
\toprule
\textbf{Sequence} & \textbf{RGB-D-Only} & \textbf{RGB-D-Inertial} \\
\midrule
LR0 & 0.014 & \bf{0.008} \\ 
LR1 & \bf{0.007} & 0.009 \\ 
LR2 & \bf{0.010} & 0.011 \\ 
LR3 & (0.118) & \bf{0.010} \\ 
\bottomrule 
\end{tabular}
\end{center}
\end{table}

Although slow, some of the sequences in the modified ICL-NUIM dataset are still difficult for dense SLAM systems to follow, particularly the last sequence.
In this sequence, the camera moves slowly along a wall providing little photometric or geometric variation.
The results of the original ElasticFusion were not obtained using the same set of internal parameters for each sequence in the ICL-NUIM dataset.
However, in this work, to showcase the robustness of our system and to avoid overfitting to a particular sequence, the default set of parameters was used across all datasets.
As a result, the RGB-D-only version of ElasticFusion now fails on this sequence.
The RGB-D-inertial system, however, is able to use the inertial data to get through the difficult section of the sequence and successfully reconstructs the scene.
The RGB-D-inertial system performs approximately as well as the RGB-D-only system on the three easier sequences.

\subsection{Real World Data}

\begin{table}[h]
\caption{Comparison of ATE RMSE on the real world datasets (brackets indicate a tracking failure)}
\label{real_ate}
\begin{center}
\begin{tabular}{lccc}
\toprule
\textbf{Trajectory Type} & \textbf{Sequence} & \textbf{RGB-D-Only} & \textbf{RGB-D-Inertial} \\
\midrule
\multirow{3}{*}{slow} & 1 & 0.227 & \bf{0.066} \\ 
& 2 & 0.110 & \bf{0.065} \\ 
& 3 & 0.225 & \bf{0.088} \\ 
\midrule 
\multirow{3}{*}{slow, loop closure} & 4 & 0.089 & \bf{0.050} \\ 
& 5 & 0.106 & \bf{0.048} \\ 
& 6 & 0.091 & \bf{0.051} \\ 
\midrule 
\multirow{3}{*}{medium} & 7 & 0.156 & \bf{0.077} \\ 
& 8 & 0.166 & \bf{0.069} \\ 
& 9 & \bf{0.118} & 0.124 \\ 
\hline 
\multirow{3}{*}{fast} & 10 & 0.098 & \bf{0.061} \\ 
& 11 & 0.438 & \bf{0.354} \\ 
& 12 & 0.267 & \bf{0.156} \\ 
\midrule 
\multirow{3}{*}{quick rotation} & 13 & 0.231 & \bf{0.110} \\ 
& 14 & \bf{0.057} & 0.063 \\ 
& 15 & 0.220 & \bf{0.064} \\ 
\midrule 
\multirow{3}{*}{low texture} & 16 & (54.238) & \bf{0.682} \\ 
& 17 & (26.306) & \bf{0.498} \\ 
& 18 & (6.536) & (2.141) \\ 
\midrule 
\multirow{3}{*}{long} & 19 & \bf{0.373} & 0.560 \\ 
& 20 & 0.359 & \bf{0.216} \\ 
& 21 & 0.417 & \bf{0.202} \\ 
\bottomrule 
\end{tabular}
\end{center}


\end{table}

While the synthetic data showed that the RGB-D-inertial system is capable of performing at least as well as the RGB-D-only system on slow, smooth trajectories, the real strength of visual-inertial systems is their robustness to aggressive motions and sequences with little photometric or geometric information.
To test this, a new dataset of 21 sequences was collected using the Intel RealSense ZR300 visual-inertial sensor.
This sensor captures aligned RGB and depth images as well as inertial measurements.
The camera intrinsics, as well as the transformation between the camera and IMU, $\T{C}{S}$, was obtained using the Kalibr calibration system \cite{Furgale:etal:IROS2013}.

To see how our system would perform under different scenarios, a number of different types of datasets were captured.
Sequences 1-3 are slow, smooth trajectories that typical
RGB-D SLAM systems could handle.
Sequences 4-6 are also slow and smooth, but with a large loop closure.
Sequences 7-9 have slightly faster trajectories, and sequences 10-12 have very aggressive trajectories but continue to map the same area, allowing the SLAM system to keep tracking against a previously built up map.
Sequences 13-15 are also aggressive trajectories, but include a quick rotation into an unmapped area of the scene.
Sequences 16-18 are slow trajectories, but pass close to a white wall such that the RGB-D data provides little photometric or geometric information. 
Sequences 19-21 are slow, smooth trajectories, but much longer than the other sequences, on the order of 15-20m.

For each sequence, a ground truth trajectory was captured using the Vicon motion capture system.
As explained in \cite{Sturm:etal:IROS2012}, these ground truth poses cannot be used to create a reliable ground truth scene reconstruction through depth image projection, as very small errors in the pose can result in very large errors in the reconstruction. For this reason, we do not calculate the reconstruction error for these sequences, only the ATE RMSE.

\begin{figure}[h!]
\centering
\includegraphics[scale=0.45]{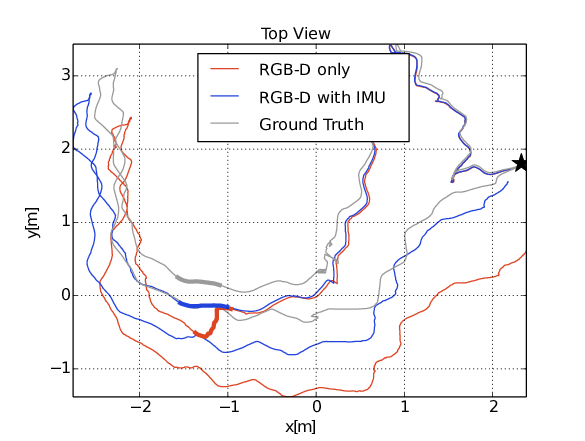}
\caption{Top view of the estimated trajectories for Sequence 21.
Integrating the IMU data improves the tracking capabilities of the framework.
In particular, it increases robustness against visually degenerate situations which pose a significant problem to the RGB-D-only framework.
Such an event, where the majority of the camera's field of view was filled with a white wall, is highlighted in the above trajectories.}
\label{topview}
\end{figure}

\subsubsection{RGB-D-Only vs. RGB-D-Inertial}

For each of the 21 sequences we compared the performance of the RGB-D-inertial system with the RGB-D-only system.
The results of these experiments are presented in Table \ref{real_ate}.
In all but 3 of the sequences the RGB-D-inertial system outperformed the RGB-D-only system, often decisively.
In particular, the RGB-D-only system was not capable of tracking the sequences where the camera moves across a white wall.
The RGB-D-inertial system is able to rely on the IMU measurements to continue tracking despite the lack of photometric or geometric information, but in the final sequence of that group, the camera moves across the wall for too long and even the RGB-D-inertial system fails.
Another example of this occurs in Sequence 21, where halfway through the trajectory the RGB-D-only system struggles when the camera goes across a blank wall. This is visualised in Fig. \ref{topview}.

\begin{figure}[h]
\centering
\includegraphics[width=0.90\linewidth]{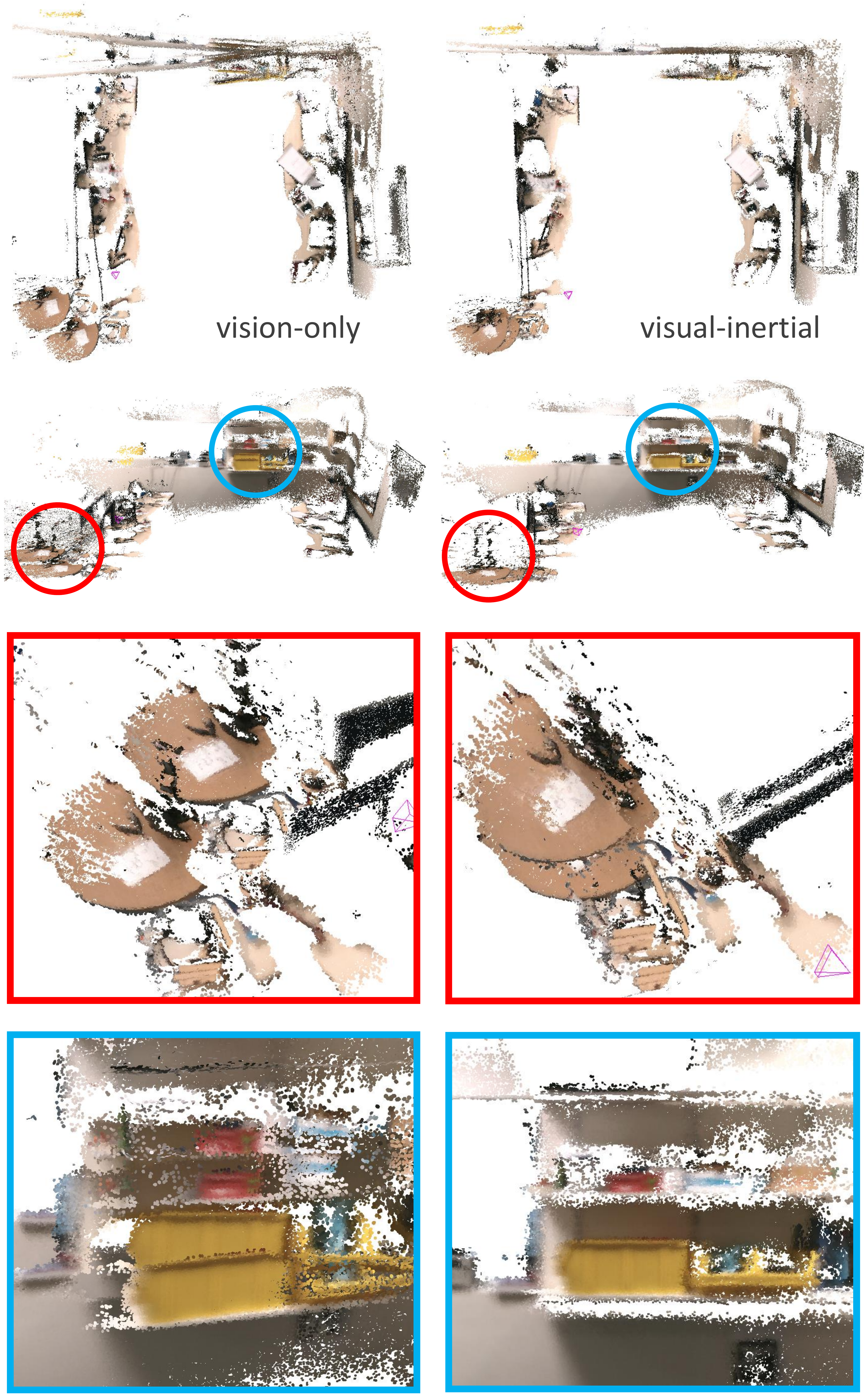}
\caption{Qualitative comparison of map reconstructions in RGB-D-only (left) and RGB-D-inertial (right) ElasticFusion: we generally achieve a higher degree of map consistency through the inclusion of inertial measurements in the tracking. While loop closure was enabled, the first zoom-in (row second from the bottom) shows that ElasticFusion failed to detect and apply a larger loop closure (in both cases); but it also shows smaller drift as a starting point before potential loop closures. The second zoom-in (bottom row) highlights in more detail the generally higher map consistency with inertial integration.}
\label{fig:visualmap}
\end{figure}

Qualitatively, we generally achieve a higher degree of map consistency in the RGB-D-inertial system compared with the RGB-D-only system. For example, Fig. \ref{fig:visualmap} shows map reconstructions when the system is run on Sequence 7, a moderately difficult sequence in the real world dataset. The top level views show how the inclusion of inertial terms in tracking significantly reduces the amount of drift, as the map is much better aligned for the RGB-D-inertial system. Keeping the map aligned helps ElasticFusion find potential loop closures, but this will still sometimes fail as shown in the pair of images second from the bottom.\footnote{We encourage the reader to view our supplementary video, available at \url{https://youtu.be/-gUdQ0cxDh0}, for a better visualisation of our qualitative results.}

\subsubsection{Odometry vs.\ SLAM}

To confirm that our formulation of a globally consistent map is improving our trajectory estimation, we examine the performance of our system with an open loop version where the system is restricted to only tracking and fusing against the active map (deformations are not allowed).
We compared this on a sample sequence from five of the different categories (we excluded the quick rotation and low texture scenes due to their difficulty). The results of these tests are shown in Table \ref{real_openLoop}.
In all cases, the closed loop version performs at least as well as the open loop version.

\begin{figure}[t]
\centering
\includegraphics[scale=0.45]{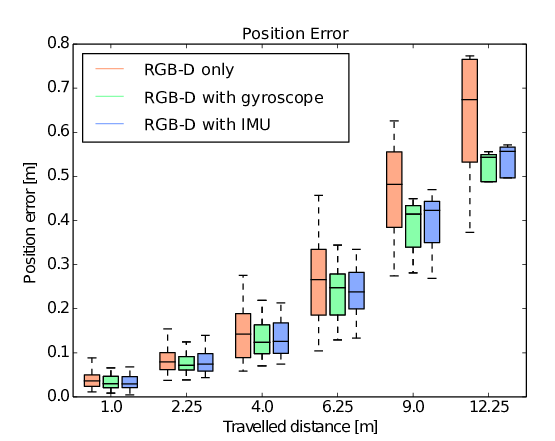}
\caption{Relation between accumulated position error and travelled distance for Sequence 20 with three different setups: no IMU, gyroscope only, and full IMU (gyroscope + accelerometer).
We observe that most of the accuracy is gained from the integration of the gyroscopes.
While the accelerometers do not significantly improve the accuracy of the system, their integration can contribute to the reliability of the system.}
\label{gyrResults}
\end{figure}

\begin{table}[h]
\caption{Comparison of ATE RMSE between open loop odometry and SLAM on the real world datasets}
\label{real_openLoop}
\begin{center}
\begin{tabular}{lccc}
\toprule
\textbf{Trajectory Type} & \textbf{Sequence} & \textbf{Odometry} & \textbf{SLAM} \\
\midrule
slow & 1 & 0.102 & \bf{0.066} \\ 
slow with loop closure & 4 & 0.051 & \bf{0.050} \\ 
medium & 7 & 0.078 & \bf{0.077} \\ 
fast & 10 & 0.062 & \bf{0.061} \\ 
long & 19 & 0.541 & \bf{0.525} \\ 
\bottomrule 
\end{tabular}
\end{center}
\end{table}

\begin{table}[h!]
\caption{Comparison of ATE RMSE between RGB-D-inertial and RGB-D with only gyroscopes on the real world datasets}
\label{real_gyro}
\begin{center}
\begin{tabular}{lccc}
\toprule
\textbf{Trajectory Type} & \textbf{Sequence} & \textbf{Gyro Only} & \textbf{Full IMU}  \\
\midrule
\multirow{3}{*}{long} & 19 & \bf{0.296} & 0.560  \\ 
& 20 & 0.223 & \bf{0.216} \\ 
& 21 & 0.203 & \bf{0.202} \\ 
\midrule 
\multirow{3}{*}{low texture} & 16 & (7.548) & \bf{0.682} \\ 
& 17 & (3.916) & \bf{0.498} \\ 
& 18 & (0.917) & (2.141) \\ 
\bottomrule 
\end{tabular}
\end{center}
\end{table}

\subsubsection{Drift Analysis}

In order to examine the relative contributions of the gyroscopes and accelerometers, we test the system on a number of sequences where the accelerometer related residuals are ignored.
Fig. \ref{gyrResults} shows the position error as a function of the distance traveled for Sequence 20, comparing RGB-D-only to RGB-D-and-gyroscopes-only to the full RGB-D-inertial system. As this figure shows, most of the gain in accuracy comes from the gyroscopes. This is confirmed by the results for the long sequences in Table \ref{real_gyro}. Over such a long sequence, the gyroscopes-only setup can outperform the full IMU due to the high noise levels of the accelerometers.
The necessity of the accelerometers, however, is shown by the low texture sequences in Table \ref{real_gyro}. As the camera passes over the white wall, the gyroscope-only system fails because without visual input the relative position is no longer constrained.


\section{CONCLUSION}
\label{s:conclusion}
We have presented what is, to the best of our knowledge, the first real-time tightly-coupled dense RGB-D-inertial SLAM system.
In the tracking step, it minimises a combined photometric, geometric and inertial energy functional to simultaneously estimate the camera pose, velocity, IMU biases and gravity direction.
In the mapping step, our system constructs a fully dense 3D reconstruction of the environment which is not only globally consistent, but gravity aligned due to the addition of an inertial deformation energy applied to the deformation graph.

We show through a series of experiments on both synthetic and real world datasets that our RGB-D-inertial system performs at least as well as the RGB-D-only version of our system on slow, smooth trajectories, but is much more robust to  aggressive motions and a lack of photometric and geometric variation.


\bibliographystyle{IEEEtran}
\bibliography{IEEEabrv,robotvision}


\end{document}